\documentclass[singlecolumn]{svjour3}
\smartqed
\usepackage[noadjust]{cite}
\usepackage{url}
\usepackage[pagebackref=true]{hyperref} 
\usepackage{xcolor}
\usepackage[most]{tcolorbox}
\usepackage{graphics} 
\usepackage{epsfig, psfig} 
\usepackage{amsmath} 
\usepackage{amssymb} 
\usepackage{morefloats}
\usepackage{subcaption}
\usepackage{bchart}
\usepackage{pgfplots}
\usepackage{array,multirow,graphicx}
\usepackage{enumitem}
\usepackage{booktabs}

\usepackage{tikz}

\begin{document}

\title{Camouflaged Object Detection and Tracking: A Survey}

\author{Ajoy~Mondal}
\institute{CVIT, International Institute of Information Technology, \\ Hyderabad, India \\email: ajoy.mondal83@gmail.com}
\maketitle

\begin{abstract}

Moving object detection and tracking have various applications, including surveillance, anomaly detection, vehicle navigation, etc. The literature on object detection and tracking is rich enough, and several essential survey papers exist. However, the research on camouflage object detection and tracking is limited due to its complexity. Existing work on this problem has been done based on either biological characteristics of the camouflaged objects or computer vision techniques. This article reviews the existing camouflaged object detection and tracking techniques using computer vision algorithms from the theoretical perspective. This article also addresses several issues of interest as well as future research direction in this area. We hope this review will help the reader learn the recent advances in camouflaged object detection and tracking.


\end{abstract}

\section{Introduction} 

Object detection is a technique which deals with detecting instances of semantic objects of a specific class (such as human, buildings, cars, etc.) in digital images and videos. It has several computer vision applications, including image retrieval, video surveillance, and some other image and video analysis tasks. The considerable success is achieved for object detection problems in a controlled environment, but the issue remains unsolved in wild places. 

\begin{figure*}
\centerline{
\tcbox[sharp corners, size = tight, boxrule=0.4mm, colframe=black, colback=white]{
\psfig{figure=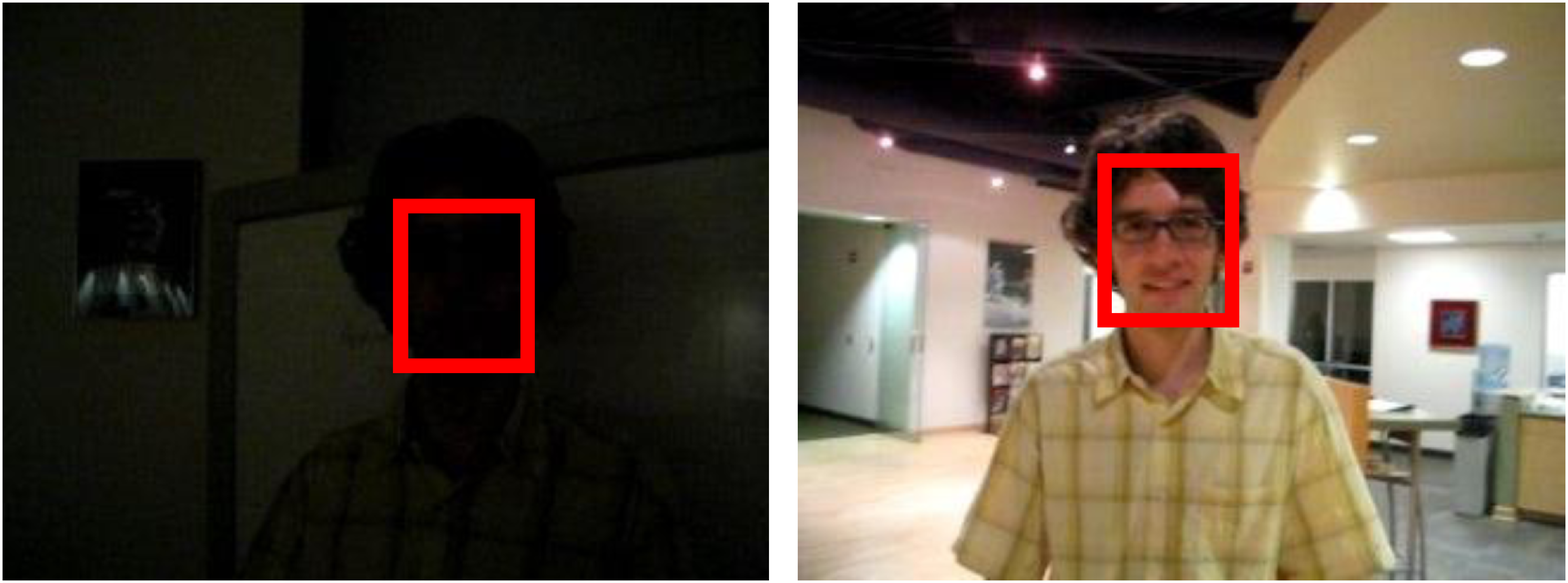, height=0.15\textwidth, width=0.45\textwidth}}
\hspace{0.0001\textwidth}
\tcbox[sharp corners, size = tight, boxrule=0.4mm, colframe=black, colback=white]{
\psfig{figure=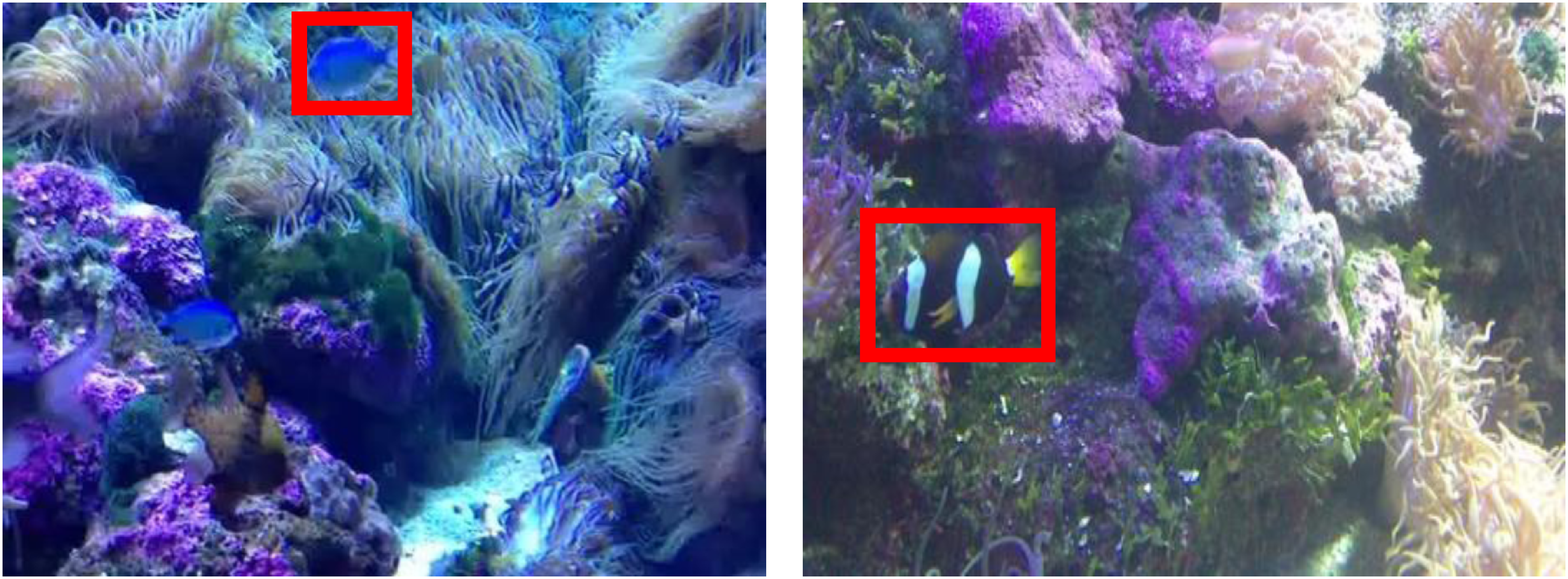, height=0.15\textwidth, width=0.45\textwidth}}}
\vspace{-0.03\textwidth}
\centerline{a \hspace{0.4\textwidth} b}
\vspace{0.003\textwidth}
\centerline{
\tcbox[sharp corners, size = tight, boxrule=0.4mm, colframe=black, colback=white]{
\psfig{figure=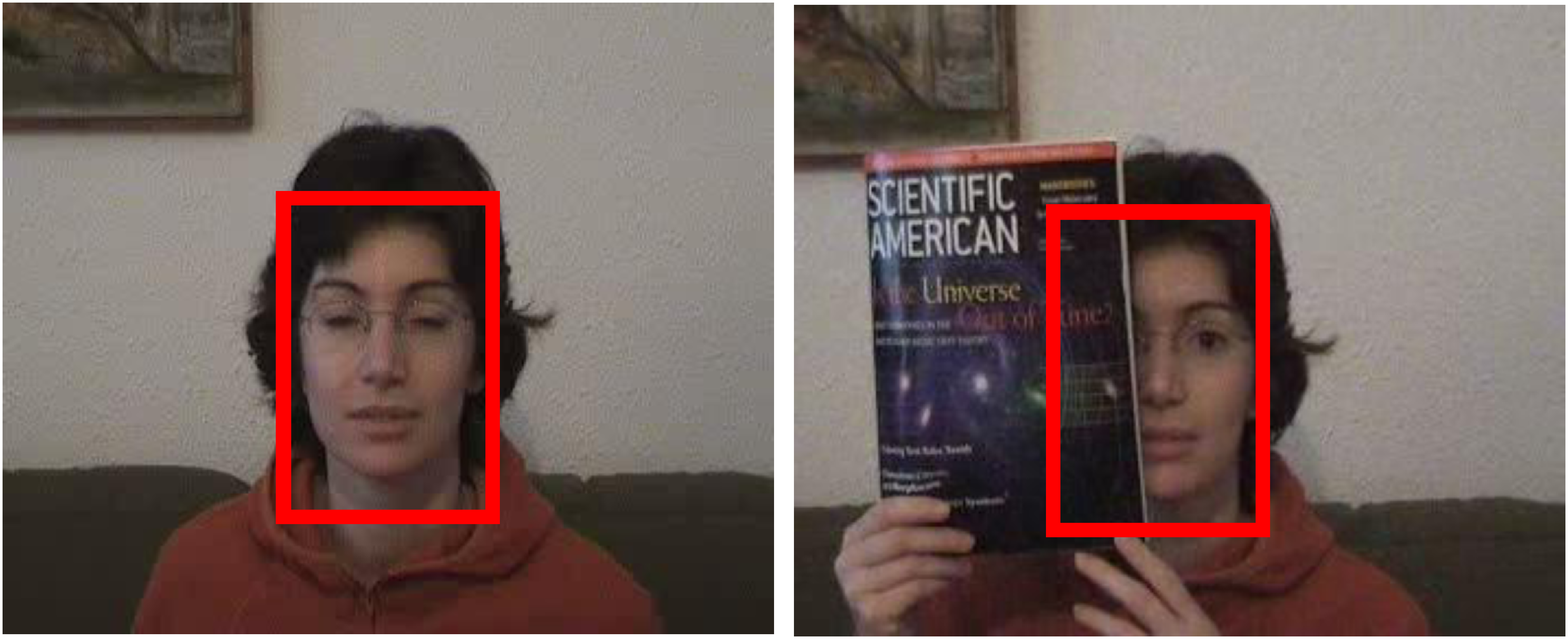, height=0.15\textwidth, width=0.45\textwidth}}
\hspace{0.0001\textwidth}
\tcbox[sharp corners, size = tight, boxrule=0.4mm, colframe=black, colback=white]{
\psfig{figure=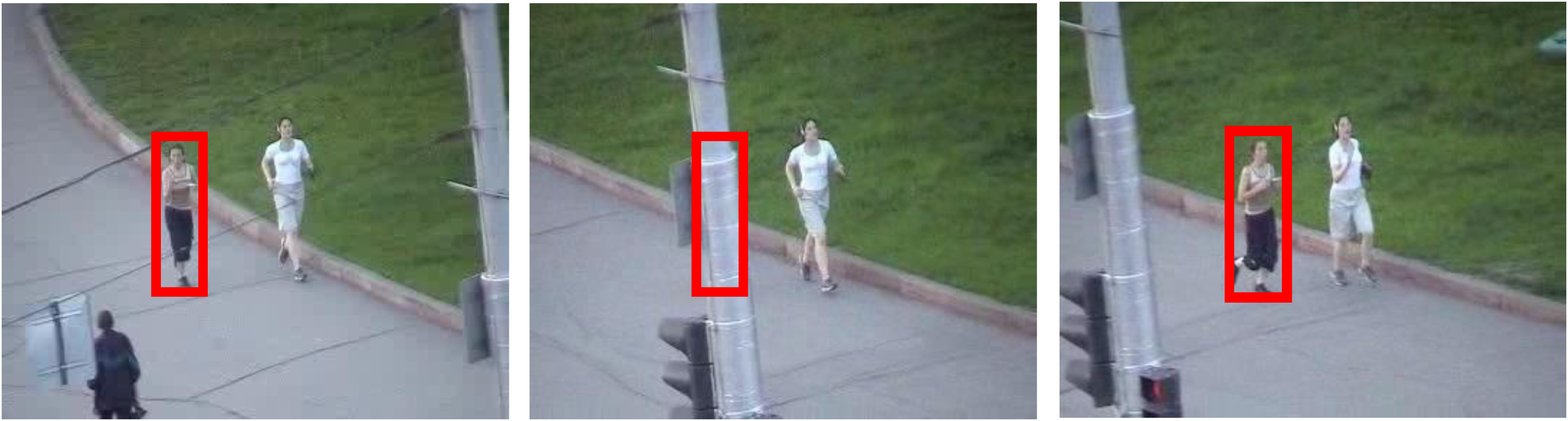, height=0.15\textwidth, width=0.45\textwidth}}}
\vspace{-0.03\textwidth}
\centerline{c \hspace{0.4\textwidth} d}
\vspace{0.003\textwidth}
\centerline{
\tcbox[sharp corners, size = tight, boxrule=0.4mm, colframe=black, colback=white]{
\psfig{figure=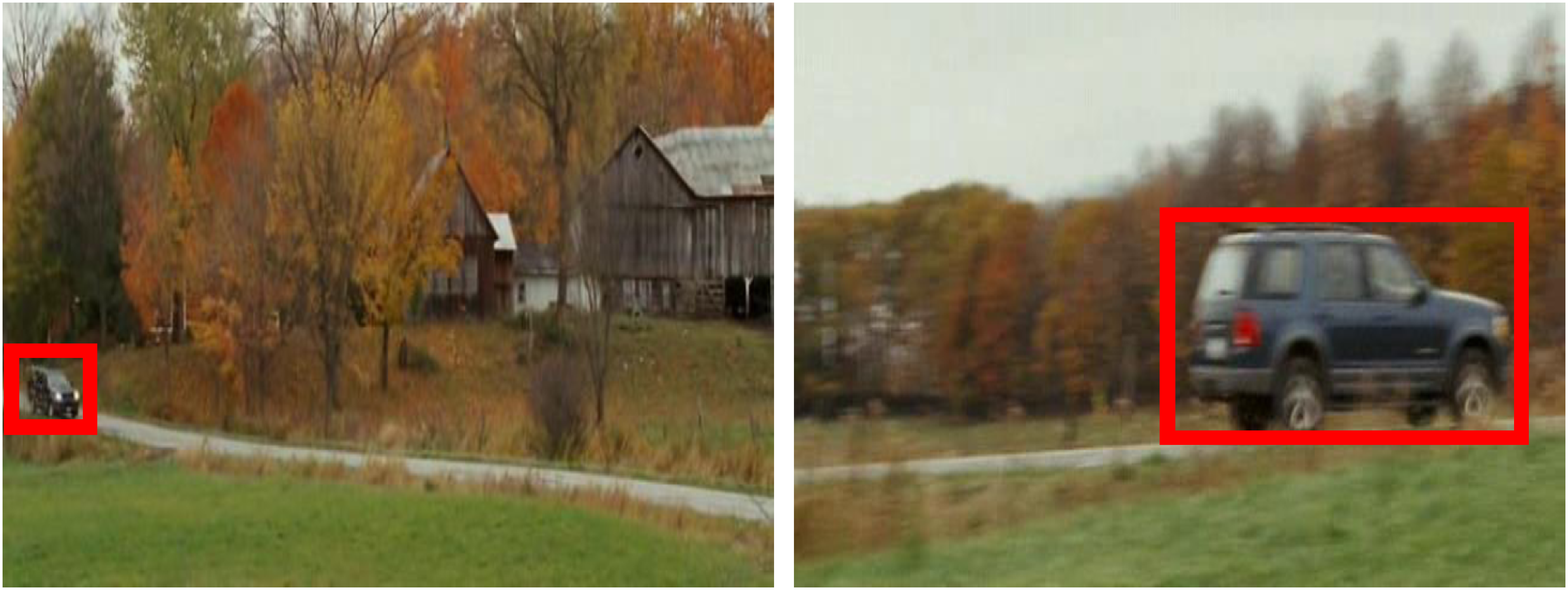, height=0.15\textwidth, width=0.45\textwidth}}
\hspace{0.0001\textwidth}
\tcbox[sharp corners, size = tight, boxrule=0.4mm, colframe=black, colback=white]{
\psfig{figure=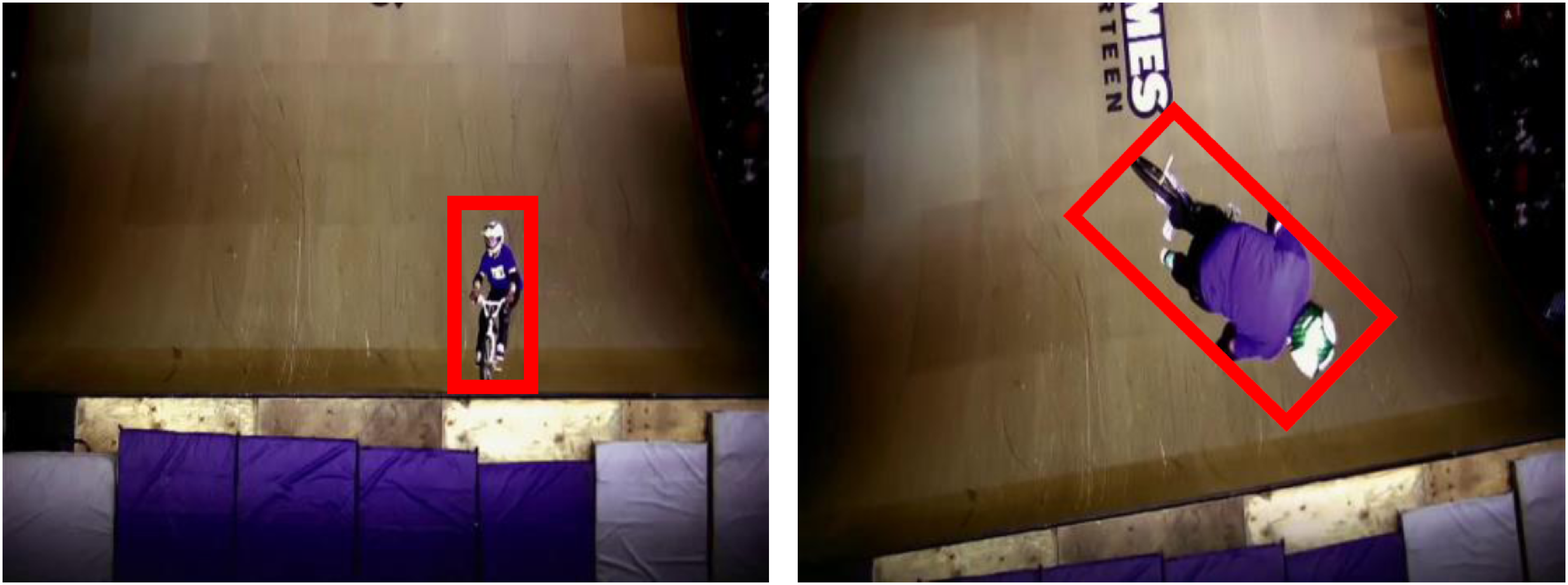, height=0.15\textwidth, width=0.45\textwidth}}}
\vspace{-0.03\textwidth}
\centerline{e \hspace{0.4\textwidth} f}
\vspace{0.003\textwidth}
\centerline{
\tcbox[sharp corners, size = tight, boxrule=0.4mm, colframe=black, colback=white]{
\psfig{figure=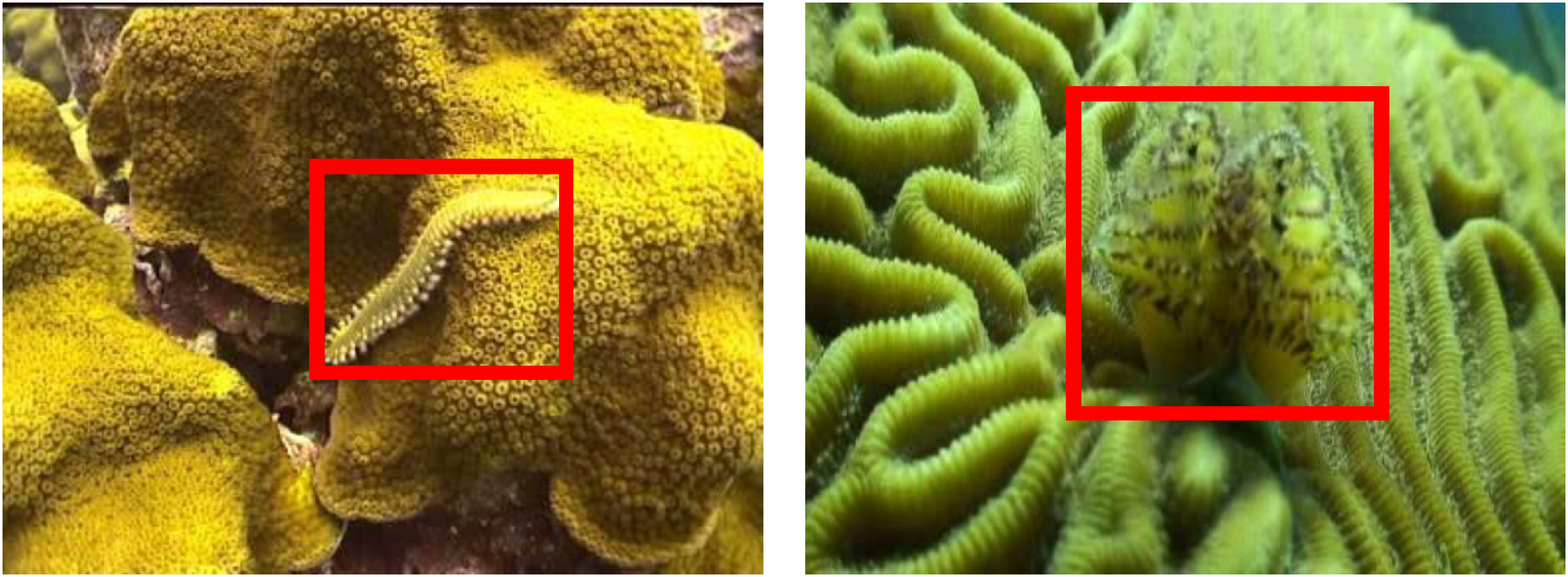, height=0.15\textwidth, width=0.45\textwidth}}
\hspace{0.0001\textwidth}
\tcbox[sharp corners, size = tight, boxrule=0.4mm, colframe=black, colback=white]{
\psfig{figure=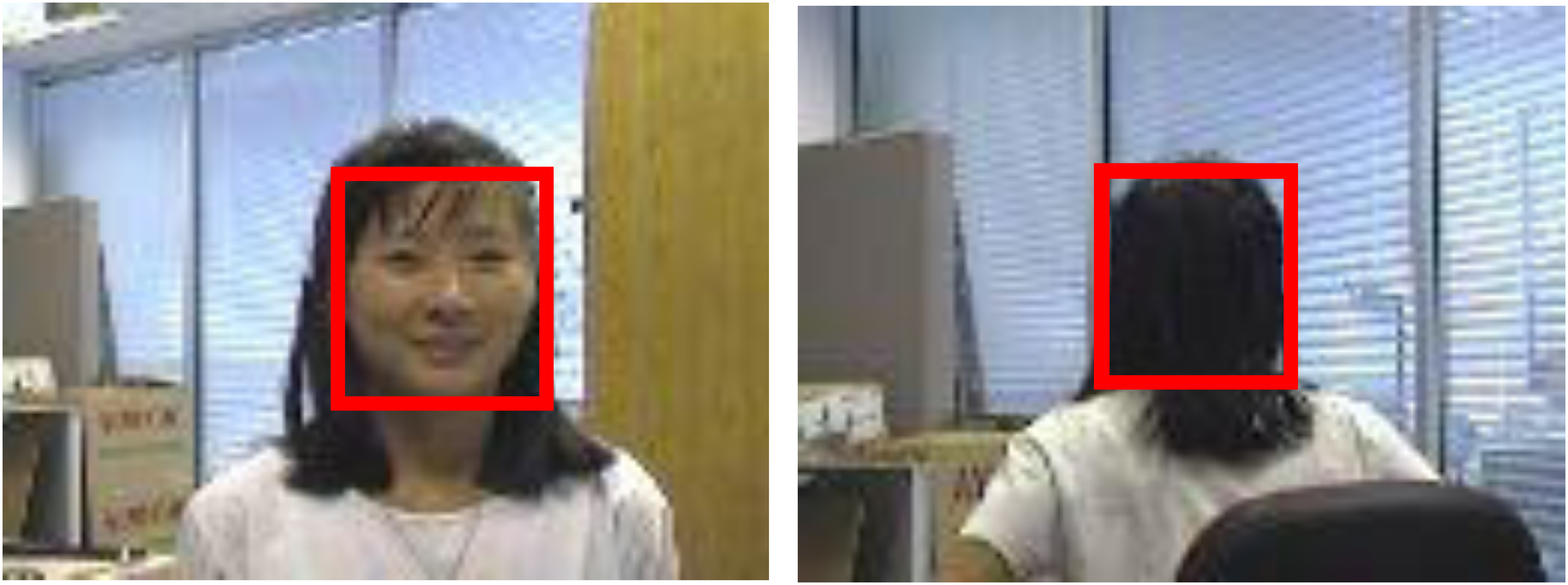, height=0.15\textwidth, width=0.45\textwidth}}}
\vspace{-0.03\textwidth}
\centerline{g \hspace{0.4\textwidth} h}
\vspace{0.003\textwidth}
\centerline{
\tcbox[sharp corners, size = tight, boxrule=0.4mm, colframe=black, colback=white]{
\psfig{figure=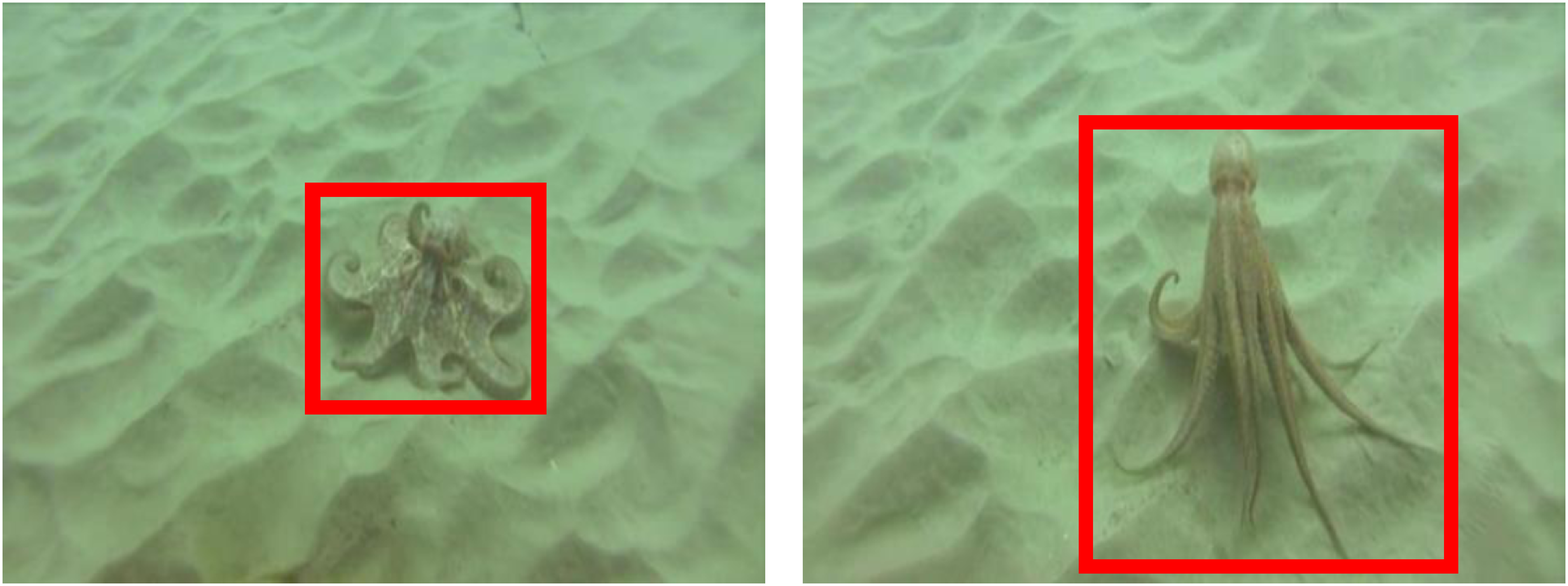, height=0.15\textwidth, width=0.45\textwidth}}}
\centerline{i}
\caption{A visual illustration of various challenges in object tracking. Examples of (a) illumination variation, (b) background clutter, (c) partial occlusion, (d) full occlusion, (e) change in object's scale, (f) change in object's orientation, (g) camouflaged object, (h) pose variation, and (i) irregular shaped object.\label{figure_chellenges}}
\end{figure*}

Visual object detection and tracking become a very challenging problem due to several factors like (i) low-quality camera sensors (including low resolution, low bit depth, low frame rate and color distortion), (ii) challenging factors (like tracking non-rigid object, tracking small object, tracking multiple objects and tracking pose varying object), (iii) requirements for real-time tracking, (iv) multi-view object tracking and (v) variations in object appearance due to several complicated factors (such as illumination variation (Figure~\ref{figure_chellenges}(a)), background clutter (Figure~\ref{figure_chellenges}(b)), partial object occlusion (Figure~\ref{figure_chellenges}(c)), full object occlusion (Figure~\ref{figure_chellenges}(d)), large variation in object scale and orientation (Figures~\ref{figure_chellenges}(e) and (f)), partially camouflaged objects (Figure~\ref{figure_chellenges} (g)), pose variation (Figure~\ref{figure_chellenges} (h)), shape deformation (Figure~\ref{figure_chellenges} (i)), rapid camera motion and noise~\cite{yilmaz2006object,smeulders2013visual,li2013survey}. Detection and (or) tracking accuracy may be degraded and even failed due to these challenges.  Numerous object tracking algorithms have been developed in the literature to handle these challenges. These invented algorithms with different properties and characteristics usually solve different visual object detection and tracking problems.

Sometimes, objects hide their signatures into their surroundings and create camouflage. The occurrence of camouflage makes object detection a more complex problem. According to Copeland and Trivedi~\cite{copeland1997models}, ``camouflage is an attempt to conceal the signature of a target into the background". In other words, camouflage is the ability of prey to hide from predators by changing their body pattern, texture, and coloration as per the environment's texture. A camouflaged object cannot be adequately visible by human vision systems. In this context, computer vision-based approaches are proposed to analyze the camouflaged objects. Work-related to camouflage can be roughly divided into two major areas: (i) camouflage assessment and design, (ii) camouflage detection, or breakin~\cite{copeland1997models,gretzmacher1998camouflage,zhangcamou}. A camouflage detection system or de-camouflaging is used to extract a target from its background. It discriminates foreground object from camouflaged image frames~\cite{singh2013survey,rong2016camouflage}. 
Camouflage detection or breaking system has many potential applications, including (i) preserving wildlife, (ii) enemy detection in the battlefield, (iii) defect detection during manufacturing, (iv) identification of duplicate products during logistics, etc~\cite{singh2013survey}. Some animals have unique biological characteristics to make them camouflaged in the environment. More research work has been done for camouflage breaking based on the biological property of these animals. The vision features of a camouflaged object are very similar to the background. The color of a camouflaged object is the same as the surrounding environment, and the texture is destroyed to merge with the background. Such characteristics of camouflaged objects make detection and tracking tasks more difficult. Due to the complexity of the problem, less work has been done using computer vision-based techniques. In this article, we review existing computer vision-based approaches for detection and tracking of a camouflaged object. We also discuss the merits and demerits of each of the algorithms. We also discuss various issues of the existing algorithms and future direction on this particular topic. We hope this review will help the reader learn the recent advances in camouflaged object detection and tracking. 

The rest of the article organizes as follows. Section~\ref{detection_camouflaged_object} discuss different existing detection and tracking algorithms for camouflaged object. Conclusive remark and future direction are presented in Section~\ref{conclusions}. 
  
\section{Camouflaged Objects Detection and Tracking} \label{detection_camouflaged_object}

Visual features of a camouflaged object are very much similar to the background --- (i) the intensity or color of the camouflaged object is close to their surrounding environment, (ii) the texture is destroyed to merge with the background, and (iii) the boundary of the camouflaged object is blurred. Such visual characteristics of camouflaged objects make detection and tracking tasks more difficult. Due to such complexity, less work has been done to attempt visual camouflage breaking in literature. However, the researchers have developed various algorithms using various visual features (e.g., intensity or color, texture, motion, gradient, etc.) to detect camouflaged objects from their surroundings. Here, we try to group the existing methods according to visual feature considering for detection and tracking of camouflaged objects. In the following subsection, the current techniques on consideration of each of the visual features are discussed.      

\subsection{Intensity/Color Features}

The feature plays an essential role in the detection of camouflaged objects. Here, techniques developed based on the intensity/color values of the frames are mainly discussed. Boult \emph{et al.}~\cite{boult2001into} developed a background subtraction technique with two thresholds to detect the camouflaged target. Here, a higher threshold value is used to detect pixels that are certainly in the foreground. The lower one is considered to detect uncertain pixels (i.e., pixels that are either part of the background or camouflaged part of the object). Then, the quasi connected component is taken into consideration to get the camouflaged target. In this case, detection accuracy is also highly dependent on thresholds. The selection of proper threshold value itself is a problem. For slow-moving objects, this method fails to detect objects.

On the other hand, Hung and Jiang~\cite{huang2005tracking} devised a method to track a camouflaged object using sequential execution of weighted region consolidation and active contour. An iterative weighted region consolidation operator is used to fill the gaps introduced by camouflage. Then, an active contour model is built during tracking to capture the actual shape of the target. The performance of this method relies on the inter-frame difference. If the object has slow motion, it is challenging to localize the object using an iterative weighted region consolidation operator. Hence, tracking may fail for sequences containing slow-moving and uniform colored objects.

In~\cite{boot2009training} Boot \textit{et al.} discussed that we usually learn something general about target recognition, which allow us to 
guide our eyes to the target more efficiently and recognize it faster and farther from fixation. They described that the background contains regular patterns. Deviations from this regularity signify the presence of a camouflaged target. However, this does not happen for all kinds of camouflaged objects. If the background and object contain similar regular patterns, it becomes challenging to extract the object from its surrounding. To detect a camouflaged object, Chandesa \emph{et al.}~\cite{chandesa2009detecting} proposed an algorithm based on particle filter. Here, the Gaussian mixture model of particle distribution is considered to investigate camouflage's effect on the particle set representing the object. This method works well on the occluded object but not for the camouflaged object. Though this method works well, it needs object information (a priori) to execute the algorithm. In~\cite{conte2009algorithm}, Conte \emph{et al.} proposed an algorithm to detect partially camouflaged people. Here, background subtraction is used to detect different parts of a person. Then grouping is performed based on a model of the shape of targets. This method is unable to provide satisfactory results for objects other than humans. In~\cite{zhang2016bayesian}, a camouflaged model is proposed using a global model for the background and integration of global and local models for the foreground. Here, both the models helped to detect camouflaged objects. 

\paragraph{Discussion:} In general, intensity or color features are elementary and computationally efficient for fast, camouflaged object detection and tracking. The intensity or color feature can detect camouflaged objects where camouflage is occurred due to texture similarity with the background. In contrast, these features cannot detect camouflaged objects where camouflage occurs due to color similarity with the environment.   

\subsection{Motion Features}

Motion is considered as an essential feature to detect an object. Several techniques have been developed based on the motion of the objects.
McKee \emph{et al.}~\cite{mckee1997stereopsis} concluded from their experiments that stereopsis is generally useful on breaking camouflage when both the observer and the scene are non-dynamic. Here, motion is a helpful feature for breaking camouflage on a static background. If the background is non-static, the motion feature fails to extract camouflaged objects from its surrounding. In this direction, Ternovskiy and Jannson~\cite{ternovskiy_1997} have proposed a motion prediction approach to detect the target in the camouflage environment. This method is suitable for the sequences where changes occur due to the object and camera movement only. If changes occur due to illumination variation, this method considers changes occurring due to object movement. Hence, this method can not work correctly in such a situation.
On the other hand, for breaking camouflage, Huimin \emph{et al.}~\cite{huimin1999} developed a computational model of visual moving image filtering in which Reichardt's elementary motion detectors~\cite{reichardt1989processing} are employed for detecting motion information. As this method relies on motion (due to object movement) information, motion due to other conditions like illumination variation, environmental condition changes, etc. produces more false alarms. All these techniques mentioned above are context-dependent and may not work well for various types of camouflaged objects.

In~\cite{hou2011detection}, Yin \emph{et al.} developed an algorithm to track a mobile object with a camouflage color based on the optical flow model.  Here, the optical flow model is used to detect motion patterns of the object and the background. The motion patterns are clustered and detect the camouflaged object based on the optical flow's magnitude and location. After that, the Kalman filter is used to improve the detection accuracy. However, the accuracy of this model depends on the results of the optical flow. For slow-moving objects and objects with camera motion, this method fails to provide excellent results.

\paragraph{Discussion:} motion plays a vital role in detecting camouflaged objects in the literature. Motion features help to detect camouflaged objects while camouflage occurs due to color/texture similarity with the background. However, motion features also fail to detect a camouflaged object with prolonged movement or stop/go motion. 

\subsection{Texture Features}

Sometimes, the object's color is similar to the background, but they have different texture patterns. The texture is considered to discriminate against the object from its surrounding. Galun \emph{et al.}~\cite{galun2003texture} developed a technique to detect camouflaged objects using a bottom-up aggregation framework that combines structural characteristics of texture elements with filter responses. It adaptively identifies the shape of texture elements and characterizes them by their size, aspect ratio, orientation, brightness, etc. Then various statistical measures of these properties are taken into account to distinguish between different textures. The said approach is applied to images containing various kinds of textures. This method works well for images containing different textures for objects and backgrounds. However, if the object and background contain a similar texture, this technique may fail to produce good results. In~\cite{bhajantri2006camouflage}, Nagabhushan and Bhajantri proposed a technique for multiple camouflage breaking using co-occurrence matrix~\cite{robert1973textural} and Canny edge detector~\cite{canny1986computational}. The co-occurrence matrix is used to analyze the given image's texture, whereas the Canny edge detector is considered to detect the edges. A combination of both co-occurrence matrix and the Canny edge detector enhances the separability between objects containing different textures. Though this method provides good results for synthetic images, it is not applied to real-life data. Also, background information needs to be known before executing this method. 

Neider and Zelinsky discussed in~\cite{neider2006searching} the detection of camouflaged targets by looking through the distracters or by scrutinizing the target-similar background. In~\cite{bhajantri2006camouflage}, Bhajantri and Nagabhushan proposed a technique to detect the camouflaged defect. Here, co-occurrence matrix-based texture features are computed within a small image region. The defective portion is detected by cluster analysis and watershed segmentation. The accuracy of this method depends on the texture feature. It may not work well for sequences where objects and background contain similar kind of texture. Sengottuvelan \emph{et al.}~\cite{sengottuvelan2008performance} developed a technique to detect the camouflaged portion of the object and extract it from the environment in a given image. Here, the grey level co-occurrence matrix ({\sc glcm}) based texture feature and dendrogram are used to detect the camouflaged object. This technique is very time-consuming due to the given image's division into several blocks or smaller regions. It does not work for images containing shading effects and object \& background containing similar textures.

Liming and Weidong~\cite{liming2010new} proposed a technique based on weighted structural similarity ({\sc wssim}) to design and evaluate camouflage texture. Here they used weighted structural similarity and original image feature to create a camouflage image. It can be used for breaking the camouflage. In~\cite{owens2014camouflaging}, Owens \textit{et al.} introduced several background matching algorithms that attempt to make the object look like whatever is behind it. It is impossible to match the background from every possible viewpoint exactly. But the proposed models are forced to make trade-offs between different perceptual factors, such as conspicuousness of occlusion boundaries and the amount of texture distortion. 

In the same direction, Li \textit{et al.} proposed a texture guided weighted voting ({\sc tgwv}) method detect foreground object in camouflaged scenes~\cite{li2017foreground}. This method employed the stationary wavelet transform to decompose the image into frequency bands. This technique could effectively capture small and hardly noticeable differences between the foreground and background in the image domain in certain wavelet frequency bands.  Finally, the foreground is detected using a weighted voting scheme based on all the wavelet bands' intensity and texture. Experimental results demonstrate that this method achieves superior performance compared to the current state-of-the-art results.

\paragraph{Discussion:} Though texture feature extraction from the color or intensity is costly. However, it is more effective for detecting camouflaged objects. While camouflage occurs due to color similarity with background, texture features give promising results in such cases.        

\subsection{Gradient Features}

When the object has a similar color, texture as the background, it is challenging to detect objects using these features. For those sequences, gradient information is useful to extract the object from the background region. In this direction, various methods have been developed to detect the camouflaged object. In~\cite{tankus1998detection,tankus2000convexity,tankus2000model,tankus2001convexity,tankus2009computer,pan2011study}, Tankus and Yeshurum proposed a $D_{arg}$ operator (context-free) to enhance an area whose shading corresponds to a convex (or concave) $3D$ object to separate such area from a flat background having similar features (like color and texture). $D_{arg}$ is applied directly to the grey-level function of the image. It responds to smooth three-dimensional convex or concave patches in objects and is not limited by any particular light source or reflectance function. Results obtained using the $D_{arg}$ operator are highly dependent on threshold values. The selection of a suitable threshold is a major issue. Further, this method does not work well for an environment containing concave background and dark-colored objects. 

\paragraph{Discussion:} Gradient features are useful in case of camouflage occur due to intensity/color and texture similarity with the background.   

\subsection{Combination of Various Features}

Camouflage occurs when the visual characteristics of the objects are too similar to the background.  The objective of a camouflage detection system is to separate the camouflaged object from the background. In such cases, features generated from single-cue (like color or texture or shape or motion) is not sufficient to extract camouflaged objects from the background because the underlying phenomenon for the occurrence of camouflage is not known. In this context, the integration of features generated from multi-cue (color, texture, shape, motion, etc.) may increase the separability between the camouflaged objects and background. Various algorithms have been proposed based on integrating different features like color, texture, motion, shape, etc.

In this direction, to detect camouflaged objects, Harville \emph{et al.}~\cite{harville2001foreground} proposed a foreground segmentation technique using both color and depth information. Information loss occurs due to a $3$D scene's projection into a $2$D picture. Here, depth information is considered to reduce the information loss due to $3$D to $2$D projection. The use of depth information increases detection accuracy and also increases the computational cost. In~\cite{kaewtrakulpong2003real}, Pong and Bowdenb proposed the use of the stochastic process to handle camouflage. This algorithm assumes that camouflage occurs when the new observation can not be associated with an existing task. The performance of this method depends on the color and motion of the object. Here, color is used to distinguish the camouflaged object and its surrounding, whereas motion is considered to separate out moving objects and the static part of the frame. Hence, both this information helps to detect moving camouflaged objects. Sometimes, this method may fail for an object, which is very similar to the background, and it contains more static objects. In~\cite{brady2003bootstrapped}, Brady and Kersten proposed a computational approach to combine low-level features with high-level models to detect and recognize an object when camouflage is present. Here, bootstrapped learning is considered to build the shape of the object, and it depends on the result of object detection. If object detection is not perfect, then recognition may not always be correct, and it fails to track correctly. 

Furthermore, in~\cite{loza2009structural}, Losa \textit{et al.} proposed a particle filter based on structural similarity measure for 
tracking camouflaged objects. Here, the structural similarity measure reflects the distance between two frames by comparing their luminance, contrast, and spatial characteristics. Here, the used measure is sensitive to relative rather than absolute changes in the frame. However, this technique may not be robust to significant alteration of the tracked object. 

However, Jiang~\cite{jiang2011object} proposed a model to track the object in a video sequence by locating a list of object features that are ranked according to their ability to differentiate against the background. Here, a mixture of color, texture, and object motion is considered multidimensional to represent the object. Multi-features representation increases the separability between the objects and their surrounding. Experimental results show that this model works well for detecting the camouflaged object. 

Recently, Mondal \textit{et al.} developed a camouflaged object tracking algorithm using a probabilistic neural network and fuzzy energy-based active contour~\cite{mondal2016partially}. Here, multi-features like color, texture, and shape are integrated to increase the discrimination between camouflaged objects and their background. The camouflaged object is detected by modifying the probabilistic neural network, and finally, it is tracked by the proposed fuzzy energy-based active contour model. The experiments showed that this method could provide good tracking results under a camouflaged environment.

\paragraph{Discussion:} Color features work while camouflage occurs due to texture similarity with the background. Texture features are effective, while camouflage occurs due to color similarity. The motion of the camouflaged object can detect it while the object with good motion. These features, e.g., color, texture, and motion, can detect the camouflaged object in a specific environment. A combination of various features, e.g., color, texture, motion, etc. effectively work without constraints.  

\subsection{Deep Features}

As the color of the camouflaged object is similar to the background, detection of it is challenging. In~\cite{li2017camouflage}, the authors considered the image enhancement technique to increase the discrimination between object and background. The region proposal network ({\sc rpn}) is considered to estimate the required target's accurate positioning. The deep neural network is used to identify extracted RoI. Finally, the detection of a camouflaged object is done. In this direction, Fang \textit{et al.}~\cite{deep_2019} proposed strong semantic dilation network to detect camouflaged people. The authors claimed that full use of semantic information in convolutional neural networks and dilated convolutions are also added to enlarge the receptive field to find camouflage people. The authors presented impressive results on a standard dataset.   

\paragraph{Discussion:} All the features, e.g., color, texture, motion, and gradient, are hand-crafted and may not use for all types of camouflaged objects. On the contrary, the in-depth feature is learned by the network from the extensive training images. The deep feature is more generic than the hand-crafted features. It works better than hand-crafted features.       

\subsection{Other Techniques}

Some other approaches that exist in the literature are discussed in this section. Marouani \emph{et al.}~\cite{marouani1995model} developed 
aircraft recognition technique in the presence of camouflage. It works on low-level matching between segments of the projection of a $3D$ model of the objects of interest. In~\cite{Guilan_1997}, Guilan and Shunqing discussed the use of spectral pattern recognition techniques to distinguish color camouflage from green vegetation background. Here, the considered spectral feature is the basic characteristic for recognizing the target. Experimentally, they concluded that the considered feature is optimum for recognition than other state-of-the-art features. However, it cannot provide good results for camouflaged objects in the background other than green vegetation. 

Beiderman \textit{et al.}~\cite{beiderman2010optical} presented a novel approach where the secondary speckle pattern is monitored for a time, to extract the temporal/spectral signature of the objects. Special image processing algorithms allow obtaining a unique signature of the object region that can be used to classify, recognize, and identify objects. This approach can detect and recognize camouflaged objects. 

Pan \textit{et al.}~\cite{pan2011study} developed an algorithm to detect the camouflaged portion of the object in a complex background with the 
help of $D_{arg}$ operator. The main drawback of this method is the selection of proper threshold value. Liu and Huang~\cite{liu2012foreground} developed a novel foreground object detection scheme that integrates the top-down information based on the expectation-maximization ({\sc em}) framework. Here, top-down information is incorporated into the object model in a generalized {\sc em} framework. A foreground model is constructed based on the object model and the state of each target. The author concluded that the method is giving good results for detecting camouflaged objects. Recently, Malathi and Bhuyan~\cite{joy2013department} developed a background subtraction scheme to detect the camouflaged object. Here, pixels corresponding to the background are quantized into codebooks, sufficient to represent a background. Codebooks extracted for each of the cameras are combined to extract the foreground from the background. To detect the camouflaged target in complex background, Zhou and Cun-chao~\cite{zhou2013cluster} proposed a novel spectral-polarimetric image fusion algorithm based on Shearlet transform. 
Kernel fuzzy c-means clustering algorithm is applied to a fused spectral-polarimetric image to separate camouflaged object from its background.

In~\cite{kim2014unsupervised}, Kim proposed a fully autonomous feature selection and camouflaged object detection method based on the online analysis of spectral and spatial features for hyper-spectral images. Here, a statistical distance metric is considered to generate candidate feature bands, and entropy-based spatial grouping property is used to reduce useless feature bands. Camouflaged objects are detected by optical spectral-spatial feature analysis with less computational complexity. Mangale and Khambete~\cite{mangale2016camouflaged} fused Thermal infrared and visible spectrum imaging modality to detect camouflaged objects. The authors presented impressive results. Li \textit{et al.}~\cite{li2018fusion} proposed a camouflaged object detection model in wavelet transformed feature space. In this method, the likelihood of each wavelet coefficient being foreground is estimated by formulating foreground and background models for each wavelet band. This method effectively aggregates the likelihoods from different wavelet bands based on the wavelet transforms' characteristics detects camouflaged objects.

\section{Conclusion and Future Directions} \label{conclusions}

In this article, a review of camouflaged object detection and tracking using computer vision-based approaches are presented. Due to the complexity of the problem, less research work has been done for breaking camouflage. Among them, very few works have been done for extracting camouflaged objects in an unsupervised way. It is crucial to detect the camouflaged object in the real scenario. As color, texture, and shape of the camouflaged object are very similar to its surroundings, techniques based on the single feature cannot extract camouflaged objects well. The integration of several features like color, texture, motion, shape, and some other features related to camouflaged objects' biological characteristics may be considered to improve the camouflage braking system. However, camouflage breaking has several crucial real-life applications, including soldier detection on the battlefield and wildlife preservation. In such cases, the unsupervised technique is necessary to detect such camouflaged objects.

\bibliographystyle{spmpsci}

\begin{thebibliography}{100}

\bibitem{copeland1997models}
A.~C. Copeland and M.~M. Trivedi, ``Models and metrics for signature strength
  evaluation of camouflaged targets,'' in \emph{SPIE proceedings series}.\hskip
  1em plus 0.5em minus 0.4em\relax Society of Photo-Optical Instrumentation
  Engineers, 1997, pp. 194--199.

\bibitem{beiderman2010optical}
Beiderman, Y., Teicher, M., Garcia, J., Mico, V., Zalevsky, Z.: Optical
  technique for classification, recognition and identification of obscured
  objects.
\newblock Optics Communications \textbf{283}(21), 4274--4282 (2010)

\bibitem{bhajantri2006camouflage}
Bhajantri, N.U., Nagabhushan, P.: Camouflage defect identification: a novel
  approach.
\newblock In: 9th IEEE International Conference on Information Technology,
  ICIT'06, pp. 145--148 (2006)

\bibitem{boot2009training}
Boot, W.R., Neider, M.B., Kramer, A.F.: Training and transfer of training in
  the search for camouflaged targets.
\newblock Attention, Perception, \& Psychophysics \textbf{71}(4), 950--963
  (2009)

\bibitem{boult2001into}
Boult, T.E., Micheals, R.J., Gao, X., Eckmann, M.: Into the woods: Visual
  surveillance of noncooperative and camouflaged targets in complex outdoor
  settings.
\newblock Proceedings of the IEEE \textbf{89}(10), 1382--1402 (2001)

\bibitem{brady2003bootstrapped}
Brady, M.J., Kersten, D.: Bootstrapped learning of novel objects.
\newblock Journal of Vision \textbf{3}(6), 413--422 (2003)

\bibitem{canny1986computational}
Canny, J.: A computational approach to edge detection.
\newblock IEEE Transactions on Pattern Analysis and Machine Intelligence
  \textbf{8}(6), 679--698 (1986)

\bibitem{chandesa2009detecting}
Chandesa, T., Pridmore, T., Bargiela, A.: Detecting occlusion and camouflage
  during visual tracking.
\newblock In: IEEE International Conference on Signal and Image Processing
  Applications (ICSIPA), pp. 468--473 (2009)

\bibitem{conte2009algorithm}
Conte, D., Foggia, P., Percannella, G., Tufano, F., Vento, M.: An algorithm for
  detection of partially camouflaged people.
\newblock In: 6th IEEE International Conference on Advanced Video and Signal
  Based Surveillance, AVSS'09, pp. 340--345 (2009)

\bibitem{copeland1997models}
Copeland, A.C., Trivedi, M.M.: Models and metrics for signature strength
  evaluation of camouflaged targets.
\newblock In: AeroSense'97, pp. 194--199 (1997)

\bibitem{galun2003texture}
Galun, M., Sharon, E., Basri, R., Brandt, A.: Texture segmentation by
  multiscale aggregation of filter responses and shape elements.
\newblock In: Proceedings of 9th IEEE International Conference on Computer
  Vision, pp. 716--723 (2003)

\bibitem{gretzmacher1998camouflage}
Gretzmacher, F.M., Ruppert, G.S., Nyberg, S.: Camouflage assessment considering
  human perception data.
\newblock In: Aerospace/Defense Sensing and Controls, pp. 58--67 (1998)

\bibitem{harville2001foreground}
Harville, M., Gordon, G., Woodfill, J.: Foreground segmentation using adaptive
  mixture models in color and depth.
\newblock In: Proceedings of IEEE Workshop on Detection and Recognition of
  Events in Video, pp. 3--11 (2001)

\bibitem{hou2011detection}
Hou, J.Y.Y.H.W., Li, J.: Detection of the mobile object with camouflage color
  under dynamic background based on optical flow.
\newblock Procedia Engineering \textbf{15}, 2201--2205 (2011)

\bibitem{huang2005tracking}
Huang, Z.Q., Jiang, Z.: Tracking camouflaged objects with weighted region
  consolidation.
\newblock In: Proceedings of Digital Image Computing: Techniques and
  Applications, DICTA'05, pp. 24--31 (2005)

\bibitem{jiang2011object}
Jiang, Z.: Object modelling and tracking in videos via multidimensional
  features.
\newblock ISRN Signal Processing \textbf{2011}, 1--15 (2011)

\bibitem{kaewtrakulpong2003real}
KaewTrakulPong, P., Bowden, R.: A real time adaptive visual surveillance system
  for tracking low-resolution colour targets in dynamically changing scenes.
\newblock Image and Vision Computing \textbf{21}(10), 913--929 (2003)

\bibitem{kim2014unsupervised}
Kim, S.: Unsupervised spectral-spatial feature selection-based camouflaged
  object detection using {VNIR} hyperspectral camera.
\newblock The Scientific World Journal \textbf{2014}, 1--8 (2014)

\bibitem{li2017camouflage}
Li, C., Zhao, X., Wan, Y.: The camouflage color target detection with deep
  networks.
\newblock In: International Conference on Neural Information Processing, pp.
  476--485. Springer (2017)

\bibitem{li2017foreground}
Li, S., Florencio, D., Zhao, Y., Cook, C., Li, W.: Foreground detection in
  camouflaged scenes.
\newblock arXiv preprint arXiv:1707.03166  (2017)

\bibitem{liming2010new}
Liming, S., Weidong, G.: A new camouflage texture evaluation method based on
  {WSSIM} and nature image features.
\newblock In: IEEE International Conference on Multimedia Technology (ICMT),
  pp. 1--4 (2010)

\bibitem{liu2012foreground}
Liu, Z., Huang, K., Tan, T.: Foreground object detection using top-down
  information based on {EM} framework.
\newblock IEEE Transactions on Image Processing \textbf{21}(9), 4204--4217
  (2012)

\bibitem{loza2009structural}
{\L}oza, A., Mihaylova, L., Bull, D., Canagarajah, N.: Structural
  similarity-based object tracking in multimodality surveillance videos.
\newblock Machine Vision and Applications \textbf{20}(2), 71--83 (2009)

\bibitem{huimin1999}
Lu, H., Wang, X., Liu, S., Shi, M., Guo, A.: The possible mechanism underlying
  visual anti-camouflage: a model and its real-time simulation.
\newblock IEEE Transactions on Systems, Man and Cybernetics, Part A: Systems
  and Humans \textbf{29}(3), 314--318 (1999)

\bibitem{joy2013department}
Malathi, T., Bhuyan, K.M.: Foreground object detection under camouflage using
  multiple camera-based codebooks.
\newblock In: Annual IEEE India Conference (INDICON), pp. 1--6 (2013)

\bibitem{marouani1995model}
Marouani, S., Huertas, A., Medioni, G.: Model-based aircraft recognition in
  perspective aerial imagery.
\newblock In: Proceedings of International Symposium on Computer Vision, pp.
  371--376 (1995)

\bibitem{mckee1997stereopsis}
McKee, S.P., Watamaniuk, S.N., Harris, J.M., Smallman, H.S., Taylor, D.G.: Is
  stereopsis effective in breaking camouflage for moving targets?
\newblock Vision Research \textbf{37}(15), 2047--2055 (1997)

\bibitem{mondal2016partially}
Mondal, A., Ghosh, S., Ghosh, A.: Partially camouflaged object tracking using
  modified probabilistic neural network and fuzzy energy based active contour.
\newblock International Journal of Computer Vision \textbf{122}(1), 1--33
  (2016)

\bibitem{neider2006searching}
Neider, M.B., Zelinsky, G.J.: Searching for camouflaged targets: Effects of
  target-background similarity on visual search.
\newblock Vision Research \textbf{46}(14), 2217--2235 (2006)

\bibitem{owens2014camouflaging}
Owens, A., Barnes, C., Flint, A., Singh, H., Freeman, W.: Camouflaging an
  object from many viewpoints.
\newblock In: IEEE Conference on Computer Vision and Pattern Recognition
  (CVPR), pp. 2782--2789 (2014)

\bibitem{pan2011study}
Pan, Y., Chen, Y., Fu, Q., Zhang, P., Xu, X.: Study on the camouflaged target
  detection method based on {3D} convexity.
\newblock Modern Applied Science \textbf{5}(4), 152--156 (2011)

\bibitem{reichardt1989processing}
Reichardt, W., Egelhaaf, M., Guo, A.K.: Processing of figure and background
  motion in the visual system of the fly.
\newblock Biological Cybernetics \textbf{61}(5), 327--345 (1989)

\bibitem{robert1973textural}
Robert, M.H., Shanmugam, K., Dinstein, I.: Textural features for image
  classification.
\newblock IEEE Transactions on Systems, Man, and Cybernetics \textbf{3}(6),
  610--621 (1973)

\bibitem{sengottuvelan2008performance}
Sengottuvelan, P., Wahi, A., Shanmugam, A.: Performance of decamouflaging
  through exploratory image analysis.
\newblock In: 1st IEEE International Conference on Emerging Trends in
  Engineering and Technology, ICETET'08, pp. 6--10 (2008)

\bibitem{singh2013survey}
Singh, S.K., Dhawale, C.A., Misra, S.: Survey of object detection methods in
  camouflaged image.
\newblock IERI Procedia \textbf{4}, 351--357 (2013)

\bibitem{Guilan_1997}
Song, G., Shunqing, T.: Method for spectral pattern recognition of color
  camouflage.
\newblock Optical Engineering \textbf{36}(6), 1779--1781 (1997)

\bibitem{tankus1998detection}
Tankus, A., Yeshurun, Y.: Detection of regions of interest and camouflage
  breaking by direct convexity estimation.
\newblock In: Proceedings of IEEE Workshop on Visual Surveillance, pp. 42--48
  (1998)

\bibitem{tankus2000convexity}
Tankus, A., Yeshurun, Y.: Convexity-based camouflage breaking.
\newblock In: Proceedings of 15th International Conference on Pattern
  Recognition, vol.~1, pp. 454--457 (2000)

\bibitem{tankus2000model}
Tankus, A., Yeshurun, Y.: A model for visual camouflage breaking.
\newblock In: Biologically Motivated Computer Vision, pp. 139--149 (2000)

\bibitem{tankus2001convexity}
Tankus, A., Yeshurun, Y.: Convexity-based visual camouflage breaking.
\newblock Computer Vision and Image Understanding \textbf{82}(3), 208--237
  (2001)

\bibitem{tankus2009computer}
Tankus, A., Yeshurun, Y.: Computer vision, camouflage breaking and
  countershading.
\newblock Philosophical Transactions of the Royal Society of London B:
  Biological Sciences \textbf{364}(1516), 529--536 (2009)

\bibitem{ternovskiy_1997}
Ternovskiy, I.V., Jannson, T.P.: Mapping-singularities-based motion estimation.
\newblock In: Optical Science, Engineering and Instrumentation'97, vol. 3173,
  pp. 317--321 (1997)

\bibitem{zhou2013cluster}
Zhou, P.C., Liu, C.C.: Camouflaged target separation by spectral-polarimetric
  imagery fusion with shearlet transform and clustering segmentation.
\newblock In: ISPDI 2013-5th International Symposium on Photoelectronic


\bibitem{smeulders2013visual}
Smeulders, Arnold WM and Chu, Dung M and Cucchiara, Rita and Calderara, Simone and Dehghan, Afshin and Shah, Mubarak: Visual tracking: An experimental survey.
\newblock IEEE Transactions on Pattern Analysis and Machine Intelligence \textbf{37}(7), 1442--1468 (2013)

\bibitem{li2013survey}
Li, Xi and Hu, Weiming and Shen, Chunhua and Zhang, Zhongfei and Dick, Anthony and Hengel, Anton Van Den: A survey of appearance models in visual object tracking.
\newblock ACM transactions on Intelligent Systems and Technology (TIST) \textbf{4}(4), 1--58 (2013)

\bibitem{yilmaz2006object}
Yilmaz, Alper and Javed, Omar and Shah, Mubarak: Object tracking: A survey. \newblock Acm computing Surveys (CSUR)\textbf{34}(4) 1--45 (2006)
  
\bibitem{deep_2019}
Zheng Fang Xiongwei Zhang Xiaotong Deng: Camouflage people detection via strong semantic dilation network. \newblock TURC-AIS, 1--7 (2019) 

\bibitem{rong2016camouflage}
Rong, Xianhui and Jia, Qi and Xu, Weidong and Lv, Xuliang and Hu, Jianghu: Camouflage Effect Evaluation of Pattern Painting Based on Moving Object Detection. \newblock International Conference on Energy, Power and Electrical Engineering 244--247 (2016)

\bibitem{zhangcamou}
Zhang, Yang and Hassan Foroosh, Philip David and Gong, Boqing: CAMOU: Learning a Vehicle Camouflage for Physical Adversarial Attack on Object Detectors in the Wild \newblock ICRL 1--20 (2019)

\bibitem{mangale2016camouflaged}
Mangale, Supriya and Khambete, Madhuri:Camouflaged target detection and tracking using thermal infrared and visible spectrum imaging. \newblock International Symposium on Intelligent Systems Technologies and Applications 193--207 (2016) 

\bibitem{zhang2016bayesian}
Zhang, Xiang and Zhu, Ce and Wang, Shuai and Liu, Yipeng and Ye, Mao: A Bayesian approach to camouflaged moving object detection. \newblock IEEE transactions on circuits and systems for video technology \textbf{27}(9) 2001--2013 (2016)
  
\bibitem{li2018fusion}
Li, Shuai and Florencio, Dinei and Li, Wanqing and Zhao, Yaqin and Cook, Chris: A fusion framework for camouflaged moving foreground detection in the wavelet domain.  \newblock IEEE Transactions on Image Processing \textbf{27}(8) 3918--3930 (2018)

\end{thebibliography}

\end{document}